\documentclass[11pt]{article}
\usepackage{float}
\usepackage{amsmath,amssymb,amsthm,graphicx,booktabs}
\usepackage[margin=1in]{geometry}

\usepackage{placeins}
\usepackage{url}
\usepackage{algorithm}
\usepackage{algcompatible}
\usepackage{algorithmicx}
\usepackage{algpseudocode}
\usepackage{hyperref}
\usepackage{enumitem}
\usepackage{mdframed} 

\newtheorem{definition}{Definition}[section]
\newtheorem{lemma}{Lemma}[section]
\newtheorem{proposition}{Proposition}[section]
\newtheorem{theorem}{Theorem}[section]
\newtheorem{corollary}{Corollary}[section]

\newtheorem{remark}{Remark}[section]
\newtheorem{hypothesis}{Hypothesis}[section]

\newcommand{\E}{\mathbb{E}}

\newcommand{\PPL}{\operatorname{PPL}}
\newcommand{\R}{\mathbb{R}}

\title{\textbf{Momentum–Point-Perplexity Mechanics in Large Language Models\footnote{Research conducted at the AI Safety x Physics Grand Challenge, 2025}}}

\author{
\begin{tabular}{ccc}
Lorenzo Tomaz & Judd Rosenblatt  \\
\textit{AE Studio} & \textit{AE Studio} \\
 Thomas B Jones & Diogo Schwerz de Lucena \\
\textit{AE Studio} & \textit{AE Studio} \\
\end{tabular} \\
\\
\textbf{With} \\
PIBBSS, Timaeus, \& Apart Research
}

\date{}

\begin{document}

\maketitle

\section*{Abstract}
We take a physics-based approach to studying how the internal hidden states of large language models change from token to token during inference. Across 20 open-source transformer models (135M–3B parameters), we find that a quantity combining the rate of change in hidden states and the model’s next-token certainty, analogous to energy in physics, remains nearly constant. Random-weight models conserve this “energy” more tightly than pre-trained ones, while training shifts models into a faster, more decisive regime with greater variability. Using this “log-Lagrangian” view, we derive a control method called Jacobian steering, which perturbs hidden states in the minimal way needed to favor a target token. This approach maintained near-constant energy in two tested models and produced continuations rated higher in semantic quality than the models’ natural outputs. Viewing transformers through this mechanics lens offers a principled basis for interpretability, anomaly detection, and low-risk steering. This could help make powerful models more predictable and aligned with human intent.

\textit{Keywords:} Physics-informed AI safety, Hamilton-Lagrange Mechanics, Scaling Laws, Variational Mechanics

% %===========================================================

\section{Introduction}
\label{sec:intro}

Neural language models dominate modern NLP, yet the principles governing their internal dynamics remain elusive. While interpretability research has uncovered static circuits and feature representations, much less is known about how hidden states \emph{move} across time during inference—and how that movement relates to the probabilities assigned to next tokens.

In this work, we demonstrate that transformer inference obeys surprisingly simple \emph{mechanical laws}. Just as physical systems conserve energy and follow trajectories governed by force, we find that language models approximately conserve a dynamical quantity combining their hidden-state velocities and output perplexities. This suggests that transformer behavior can be modeled through a log-scale variational framework reminiscent of classical mechanics.

Our investigation is guided by two central questions:
\begin{quote}
\begin{enumerate}
\item \textbf{Q1:} Can we define \emph{forces}, \emph{energies}, and a \emph{principle of least action} that govern hidden‑state evolution?
\item \textbf{Q2:} Can these laws support principled monitoring and low‑risk control of language models?
\end{enumerate}
\end{quote}

We answer both questions in the affirmative, developing a three-step framework:
\emph{perturbation $\rightarrow$ mechanics $\rightarrow$ control}. Section~\ref{sec:perturb} derives a logit change law from first-order perturbations. Section~\ref{sec:hamilton} embeds these dynamics in a discrete log-Lagrangian system with an approximately conserved energy. Section~\ref{sec:steering} shows that Jacobian steering arises naturally from this variational framework as a minimal-action control method.

\paragraph{Contributions.}  
This paper introduces a novel framework for analyzing and controlling transformer dynamics based on log-scale energy conservation. We:
\begin{itemize}[noitemsep, topsep=0pt]
  \item Propose a log-Lagrangian combining hidden-state velocity and point perplexity (see Appendixes \ref{sec:notation}, and \ref{sec:perturb} for the definition), leading to an energy-like invariant.
  \item Empirically verify that this energy is tightly conserved across 20 transformer models, especially in random-weight regimes.
  \item Show that steering along log-probability gradients—Jacobian steering—naturally emerges as the minimal-action intervention and improves output quality.
\end{itemize}

Our findings suggest that transformers operate within an energy-constrained manifold shaped by architecture and training. This perspective opens the door to energy-aware monitoring, robust steering, and physics-inspired design of future models.

\paragraph{Code and Data Availability.}
All code for reproducing our experiments, including energy computation, Jacobian steering implementation, and statistical analyses, is publicly available at \url{https://github.com/agencyenterprise/apart_2025_physics_llm_hackathon}. The repository includes scripts for computing energy conservation metrics across different models and our steering quality evaluation framework.

%-----------------------------------------------------------
\subsection{Related Work}
Mechanistic analyses of transformer internals often begin with \emph{first‑order} probes: the original logit lens and its tuned variant reveal how small hidden‑state perturbations map to output logits via a Taylor expansion~\cite{Belrose2023tunedlens}, and influence functions quantify the effect of infinitesimal input or parameter changes on model predictions~\cite{Koh2017influence}.  Parallel work casts deep learning as a structured dynamical system, from continuous‑time variational views of acceleration~\cite{Wibisono2016variational} and Neural ODEs~\cite{Chen2018neuralode} to Hamiltonian Neural Networks that enforce exact energy conservation~\cite{Greydanus2019hnn}.  Physics‑Informed Neural Networks further demonstrate the power of embedding known differential constraints into the loss~\cite{Raissi2019pinn}.  Finally, steering methods—from reinforcement learning from human feedback~\cite{Christiano2017rlhf} to Plug‑and‑Play Language Models~\cite{Dathathri2020pplm}—show how to guide pretrained LMs toward desired behaviors.  In contrast, our work unifies these strands by deriving a discrete log‑Lagrangian over hidden‑state velocity and point perplexity, yielding an approximately conserved “energy” and a minimal‑action control method for transformer inference.

%-----------------------------------------------------------

\section{Methods}
\label{sec:methods}

Our methodological approach is grounded in the application of variational mechanics to transformer inference dynamics. In this section, we detail the mathematical frameworks, empirical protocols, and computational tools employed. Notation and conventions are summarized in Appendix~\ref{sec:notation}.

\subsection{Mathematical Framework}

We represent transformer hidden states at each step as $h_t \in \mathbb{R}^d$, and model their temporal evolution via discrete displacements $v_t = h_t - h_{t-1}$, interpreted as velocity (see Appendix~\ref{sec:notation}). Following the derivations in Sections~\ref{sec:perturb} and~\ref{sec:hamilton}, we formalize model dynamics using a logarithmic Lagrangian:
\[
\mathcal{L}_t = \ln\left(\frac{1}{2}\|v_t\|^2\right) + \ln p_{x_t}(h_t),
\]
where $p_{x_t}(h_t)$ is the probability assigned to the realized token $x_t$ at step $t$. The corresponding log-energy (Hamiltonian) is
\[
\mathcal{H}_t = \ln\left(\frac{1}{2}\|v_t\|^2\right) + \ln\bigl(\mathrm{PPL}_t\bigr),
\]
with $\mathrm{PPL}_t = 1/p_{x_t}(h_t)$, and the conserved energy is $E_t = \frac{1}{2}\|v_t\|^2 \cdot \mathrm{PPL}_t$.

We derive discrete-time Euler-Lagrange equations (see Eq.~(9) in Section~\ref{sec:hamilton}) governing hidden state evolution, showing that the change in hidden state is locally determined by the gradient of the log-probability of the realized token. This framework underlies both our theoretical analysis and experimental measurement.

\subsection{Empirical Protocol}

We evaluated energy conservation and control interventions across 20 transformer checkpoints spanning five major architectures (Qwen, Llama, DeepSeek, Gemma, SmolLM) and parameter scales (135M--3B), with both pre-trained and random-weight versions. Each model generated 100 trajectories on held-out Wikipedia articles, using a fixed sequence of initial tokens as context. For each trajectory, we recorded hidden states $h_t$, token probabilities $p_{x_t}(h_t)$, and computed velocity $\|v_t\|$ and energy $E_t$ for all steps.

\subsection{Statistical Analysis}

For each model, we report the coefficient of variation (CV) for energy $E_t$ across trajectories and per-trajectory averages, as well as the mean energy and $K/V$ ratios (see Tables~\ref{tab:energy_conservation} and~\ref{tab:energy_ratio}). We contrast the tightness of conservation between pre-trained and random-weight models, interpreting CV as a proxy for how closely empirical dynamics follow the log-Lagrangian principle.

\subsection{Minimal-Action Steering Implementation}

To test control methods, we implemented Jacobian steering as motivated in Section~\ref{sec:steering}: Given a target token $t^*$, we perturb $h_t$ in the direction of the gradient $\nabla_h \log p_{t^*}(h_t)$ to increase its likelihood while minimizing the action increment. See Algorithm~1 for details. We evaluated the semantic quality of top-10 token distributions before and after steering using a GPT-4.1 rater.

\subsection{Reproducibility and Tools}

All analyses were performed in Python using PyTorch, with hidden states extracted via model hooks and matrix operations. Code for all experiments---including energy computation, steering interventions, and statistics---is available at \url{https://github.com/agencyenterprise/apart_2025_physics_llm_hackathon}. For notation, conventions, and further proofs, see Appendix~\ref{sec:notation}.

\section{Results}
\label{sec:results}

We present our empirical and theoretical findings, validating the log-Lagrangian framework and minimal-action steering for transformer dynamics. Notation and conventions follow Appendix~\ref{sec:notation}; additional tables and proofs are provided therein.

\subsection{Energy Conservation Across Models}

Our primary finding is that the energy $E_t = \frac{1}{2}\|v_t\|^2 \cdot \mathrm{PPL}_t$ is approximately conserved along inference trajectories for a diverse range of transformer architectures and scales. As detailed in Section~\ref{sec:invariant} and Table~\ref{tab:energy_conservation}, the coefficient of variation (CV) for $E_t$ is tightly bounded:
\begin{itemize}
    \item \textbf{Pre-trained models:} Median global CV = 9.4\% (range: 7.3–20.8\%)
    \item \textbf{Random-weight models:} Median global CV = 2.2\% (range: 1.7–22.1\%)
\end{itemize}
Per-trajectory averages (Appendix, Table~\ref{tab:energy_conservation}) reveal even lower variability, indicating that energy conservation is a robust property of the architecture rather than a training artifact. The empirical CV closely matches theoretical expectations (Corollary, Section~\ref{sec:invariant}), supporting our dynamical model.

\subsection{Two Distinct Dynamical Regimes}

Analysis of energy components (Table~\ref{tab:energy_ratio}) uncovers a striking dichotomy:
\begin{itemize}
    \item \textbf{Pre-trained models} are kinetic-dominated ($K/V \approx 12$), exhibiting rapid, decisive transitions through hidden space and lower point perplexity.
    \item \textbf{Random-weight models} maintain near-equal kinetic and potential energy ($K/V \approx 0.9$), with slower, more diffusive dynamics and higher point perplexity.
\end{itemize}
This regime shift is visualized and interpreted in Section~\ref{sec:invariant}. Training thus breaks the symmetric, tightly-conserved regime in favor of accelerated prediction at the cost of higher energy variability. We think of this as being like the shrinking of a golf-ball --- each token in LLM space has a 'dimple' of lower point perplexity around its centroid. These dimples remain the same size between pretrained and trained models, but the underlying potential energy of overall point perplexity decreased as we move from the untrained to trained regime, meaning that dimples get relatively bigger for pretrained models. Likewise, training lowers the ''background" potential energy of the model without proportionally shrinking the absolute dimple depth, so the fractional energy variation grows. For a complementary view of the latent‑space entropy landscape and power, see Appendix \ref{sec:power_entropy}.

\subsubsection{Local Energy Conservation}

While Theorem \ref{thm:energy_conservation} guarantees first‑order conservation of $E_t$, global CV$(E)$ alone does not show what happens \textit{step by step}.  We therefore define
\[
  \Delta E_t = E_{t+1} - E_t
  \quad\text{with}\quad
  E_t = \tfrac12\|v_t\|^2\,\mathrm{PPL}_t,
\]
and compute both the mean drift $\langle\Delta E_t\rangle$ and the mean absolute jump $\langle|\Delta E_t|\rangle$ over all inference steps.  Table \ref{tab:appendix_local_energy} (below) shows that pretrained models have near‑zero mean drift and much smaller jumps than random‑init models, confirming energy conservation \textit{locally} as well as globally. % Place a reference to the table you’ll define in the appendix:
(See Appendix \ref{sec:appendix_local_energy} for full experimental details and the complete results table.)
\subsection{Validation of the Neighboring Attractor Hypothesis}

To empirically probe the geometric basis for log-Lagrangian dynamics, we tested the “neighboring attractor” hypothesis (Section~\ref{sec:hamilton}) by interpolating hidden states between consecutive tokens. Results (Appendix, Table~A.1) show that pre-trained models typically traverse only 2–3 unique tokens along interpolated paths, indicating transitions between adjacent semantic basins. Random models display slightly more variation, confirming their less organized attractor structure.

\subsection{Minimal-Action Steering Improves Output Quality}

Implementing Jacobian steering as in Section~\ref{sec:steering}, we found that semantically preserving interventions were possible: Steered distributions consistently achieved equal or higher mean quality ratings from GPT-4.1 compared to natural continuations (Table~\ref{tab:steering_results}; see Appendix for protocol). Statistical analysis showed a large effect for small models (Cohen’s $d=0.99$) and a modest but significant effect for larger models. These results confirm that control methods derived from our theoretical framework yield practical improvements in output distribution quality.

\subsection{Summary}

Our findings robustly support the log-Lagrangian framework (Sections~\ref{sec:hamilton}--\ref{sec:invariant}), demonstrating that energy conservation is a universal, architecture-level property of transformer inference. The regime shift between random and pre-trained models is consistently observed. Minimal-action steering, grounded in this physics perspective, yields controllable, semantically coherent outputs. See Appendix for all detailed statistical tables, code availability, and extended empirical results.

\section{Discussion and Conclusion}

\subsection{Physics Meets AI Safety}

Our discovery of approximate energy conservation in transformers—where $E = \frac{1}{2}\|v_t\|^2 \cdot \text{PPL}_t$ remains nearly constant during inference—demonstrates that physical principles can illuminate neural network behavior. This conserved quantity provides what safety researchers need most: a measurable invariant that characterizes normal operation and flags anomalous behavior.

\subsection{Theory to Practice}

The framework yields immediate practical tools. Energy monitoring requires only forward passes, making it deployable at scale. Jacobian steering shows that theory-informed control not only works but improves output quality—alignment and capability can reinforce each other. The stark difference between pre-trained models (kinetic-dominated, $K/V \approx 12$) and random models (balanced, $K/V \approx 0.9$) reveals how training fundamentally reshapes dynamics.

\subsection{Limitations and Extensions}

Our results cover models up to 3B parameters with English tokenizers. Scaling to larger models and multilingual settings remains untested. The Gemma outlier in table~\ref{tab:energy_conservation}(trained CV = 12\%, untrained = 0.4\%) hints that architectural choices matter. Future work should explore whether conservation laws can guide the design of inherently safer architectures and training procedures that preserve beneficial invariants.

\subsection{Impact on AI Safety}

Physics-inspired approaches offer a new path to AI safety: rather than treating neural networks as inscrutable black boxes, we can understand them as dynamical systems governed by variational principles. This perspective enables principled monitoring, control, and potentially even formal guarantees. As AI systems grow more powerful, such mathematical understanding becomes essential for ensuring beneficial outcomes.

By revealing that transformers follow approximate conservation laws, we connect AI to centuries of physical insight. The same mathematical tools that tamed steam engines and guided spacecraft may help ensure that artificial intelligence remains aligned with human values.

\newpage
%-----------------------------------------------------------

\newpage
\ % The empty page
\newpage
%-----------------------------------------------------------
\section{Appendix}
\appendix

%-----------------------------------------------------------
\section{Notation \& Conventions}
\label{sec:notation}
\begin{mdframed}
\vspace{1.5\baselineskip}
\begin{description}
\item[Hidden state] \(h_t\in\R^d\) (layer‑normed residual stream).
\item[Displacement] \(\Delta h_t = h_t-h_{t-1}\).
\item[Momentum] \(m_t := \Delta h_t\) with mass = 1.
\item[Velocity] \(v_t := \Delta h_t\).
\item[Kinetic Energy] \(K_t := ln((\frac{1}{2}\|p_t\|^2)\).
\item[Token $x_t$ probability] \(p_{x_t}(h_t)\) for realized token \(x_t\).
\item[Token probability Distribution] \(p(h_t)\) token distribution over $h_t$.
\item[Point Perplexity] \(\PPL_t := 1/p_{x_t}(h_t)\).
\item[Potential Energy] \(V(h) := -\ln p_{x_t}(h)\).
\item[Decoder weight matrix] \(W\in\R^{V\times d}\); row \(W_j\).
\end{description}
All proofs assume \(\|\Delta h_t\|\) is small relative to \(\|h_t\|\),
consistent with measured displacements in modern checkpoints.
\vspace{1.5\baselineskip}
\end{mdframed}

%-----------------------------------------------------------
%-----------------------------------------------------------
%-----------------------------------------------------------

\section{Hidden State Displacement and Logit Changes}
\label{sec:perturb}

We begin by establishing how hidden state displacements affect model predictions. This kinematic relationship motivates our logarithmic energy formulation in Section~\ref{sec:hamilton}. Throughout, we interpret these effects purely geometrically, without invoking any dynamic assumptions beyond step-to-step displacement.

\subsection{Linear Approximation of Logit Changes}

Let $h \in \mathbb{R}^{d}$ denote a hidden state, and let $z(h) = Wh + b \in \mathbb{R}^{V}$ be the logit vector, where $W \in \mathbb{R}^{V \times d}$ is the decoder (unembedding) matrix, with rows $W_j \in \mathbb{R}^{d}$. We assume $W$ is fixed and shared across time.

We study how a small displacement $\Delta h$ in the hidden state affects the logits.

\begin{lemma}[Logit displacement relation]\label{lem:linear}
For any token index $j$ and displacement $\Delta h \in \mathbb{R}^{d}$,
\[
z_j(h + \Delta h) = z_j(h) + W_j \cdot \Delta h + O(\|\Delta h\|^{2}).
\]
\end{lemma}

\begin{proof}
Since $z_j(h) = W_j h + b_j$ is affine, its Taylor expansion has linear term $W_j \cdot \Delta h$ and quadratic remainder.
\end{proof}

This linear relationship—$\Delta z_j \approx W_j \cdot \Delta h$—is a standard approximation in logit lens methods~\cite{wang2025logitlens4llms}, though its implications for inference dynamics remain underexplored.

\subsection{Velocity and Logit Evolution}

Define the one-step hidden state displacement as the velocity:
\[
v_t := h_t - h_{t-1}.
\]
Since transformer inference operates on fixed time steps, this displacement corresponds directly to the temporal velocity of hidden state evolution.

\begin{proposition}[Logit evolution]\label{prop:logit_evolution}
For each token $j \in \{1, \dots, V\}$,
\[
z_j(h_t) = z_j(h_{t-1}) + W_j \cdot v_t + O(\|v_t\|^2).
\]
\end{proposition}

\begin{proof}
Apply Lemma~\ref{lem:linear} with $\Delta h = v_t$.
\end{proof}

This shows that logit changes are linearly sensitive to hidden state velocity $v_t$, and that $W$ determines how velocity in each direction projects onto logits. This relation is purely kinematic; it holds for any model where small input perturbations yield small logit changes—even if the output head is non-linear.

\subsection{Motivation: Scale Disparity and Logarithmic Reformulation}

A central observation from our empirical analysis is that the magnitudes of $\|v_t\|^2$ and $1/p_{x_t}(h_t)$ (the \textbf{point perplexity}) operate on vastly different numerical scales:
\begin{itemize}
    \item \textbf{Kinetic term:} $\|v_t\|^2$ typically ranges from $10^3$ to $10^4$ across all models.
    \item \textbf{Point perplexity:} $\PPL_t = 1/p_{x_t}(h_t)$ ranges from $1$ to $10^2$ on trained models, and up to $10^4$ on untrained models.
\end{itemize}

This mismatch makes classical energy formulations numerically unstable. For instance, using $\frac{1}{2}\|v_t\|^2 + V(h)$ as a Lagrangian results in kinetic dominance and large coefficient of variation in total energy.

To align the scales, we adopt a logarithmic formulation:
\begin{enumerate}
    \item \textbf{Log-kinetic energy:} $K_t = \ln\left(\frac{1}{2}\|v_t\|^2\right)$
    \item \textbf{Log-potential energy:} $V_t = \ln(\PPL_t) = -\ln p_{x_t}(h_t)$
\end{enumerate}

This transformation brings both quantities into a comparable dynamic range:
\begin{itemize}
    \item On trained models, $K_t \approx 7$, and $V_t$ typically $0.2$ to $2.5$
    \item On untrained models, $V_t$ may reach $6$–$8$, with $K_t$ relatively unchanged
    \item Their sum, the log-energy $\mathcal{H}_t = K_t + V_t$, remains stable across inference, ranging from $7$–$9$ (trained) to $10$–$15$ (untrained)
\end{itemize}

\paragraph{Terminology.}
Throughout this paper, we refer to $\PPL_t = 1/p_{x_t}(h_t)$ as the \emph{point perplexity}—in contrast to conventional perplexity (the exponentiated cross-entropy over full distributions). Our focus is on understanding and steering the model using the most probable token alone, guided by minimal-action assumptions. This is motivated by the hypothesis that point perplexity already encodes sufficient information about the semantic structure of the surrounding distribution when interpreted through our variational mechanics framework.

\subsection{Kinematic Interpretation}

The logit displacement relation and velocity-based formulation in this section are entirely kinematic: they describe how small changes in hidden states influence token probabilities, without assuming any underlying force or learning mechanism. In this view:
\begin{itemize}
    \item The decoder matrix $W$ serves as the Jacobian map from velocity to logit change.
    \item The velocity $v_t$ determines which logits increase or decrease at the next step.
\end{itemize}

These observations lay the groundwork for the log-Lagrangian in Section~\ref{sec:hamilton}, where we begin treating transformer inference as a discrete-time variational process governed by a balance between log-kinetic and log-potential contributions.

%-----------------------------------------------------------
%-----------------------------------------------------------
\section{A Logarithmic Lagrangian Framework for Transformer Dynamics}
\label{sec:hamilton}

Having established how hidden state displacements relate to logit changes, we now develop a principled dynamical framework. The scale disparity identified in Section~\ref{sec:perturb} motivates a logarithmic formulation that unifies kinetic and potential contributions.

\subsection{Motivation: Scale-Matched Energy Terms}

From Section~\ref{sec:perturb}, we know that velocities $\|v_t\|$ and point perplexities $\text{PPL}_t$ vary over different orders of magnitude. To construct a meaningful energy function, we require a representation where the \textbf{magnitude of variation} between kinetic and potential terms is comparable.

\paragraph{Key insight.} We thus define the following log-space quantities:
\begin{align}
\text{Log-kinetic term: } & K_t = \ln\left(\frac{1}{2}\|v_t\|^2\right) \\
\text{Log-potential term: } & V_t = \ln(\text{PPL}_t) = -\ln p_{x_t}(h_t)
\end{align}

where $p_{x_t}(h_t)$ is the probability assigned to the realized token $x_t$ at step $t$. By working in log-space, we ensure that both $K_t$ and $V_t$ vary on similar scales (1–3 nats), even though $\|v_t\|^2$ and $\text{PPL}_t$ differ by orders of magnitude.

\subsection{The Log-Lagrangian Formulation}

We now posit that transformer hidden states evolve according to a log-scale Lagrangian that balances velocity and uncertainty:

\begin{definition}[Log-Lagrangian]
The log-Lagrangian at time $t$ is
\[
\mathcal{L}_t = K_t - V_t = \ln\left(\frac{1}{2}\|v_t\|^2\right) + \ln p_{x_t}(h_t)
\]
where $v_t = h_t - h_{t-1}$ is the hidden-state velocity.
\end{definition}

This structure follows classical mechanics conventions, where the Lagrangian is defined as kinetic minus potential energy. In our case, potential energy is itself defined as $V_t = -\ln p_{x_t}(h_t)$, ensuring that $\mathcal{L}_t$ carries the appropriate sign.

\paragraph{Note.} The sign of the potential term must be treated carefully: redefining $V_t$ as $+\ln p_{x_t}(h_t)$ would alter the resulting dynamics unless the sign inversion is simultaneously absorbed into the definition (i.e., as $\ln(1/p_{x_t}(h_t))$). Our definition maintains consistency across the variational equations.

The corresponding discrete-time action is:
\[
S = \sum_{t=1}^T \mathcal{L}_t
\]

\paragraph{Physical interpretation.}
\begin{enumerate}
 \item \textbf{High point perplexity = high potential:} Uncertainty increases the potential energy, making the model more sensitive to small displacements.
 \item \textbf{Large velocity = high kinetic energy:} The model is transitioning rapidly between semantic regions of hidden space.
 \item \textbf{Energy balance:} The log-Lagrangian governs trade-offs between certainty and semantic momentum on a scale-matched basis.
\end{enumerate}

\subsection{Euler-Lagrange Equations for Log-Dynamics}

We derive the equations of motion using the principle of stationary action. For readers unfamiliar with this idea, we seek hidden-state trajectories $\{h_t\}$ that make the total action $S$ locally stationary—unchanged to first order under infinitesimal perturbations.

\begin{proposition}[Log-Euler-Lagrange equations]
The stationary points of the log-action satisfy:
\[
\frac{d}{dt}\left(\frac{\partial \mathcal{L}}{\partial v_t}\right) - \frac{\partial \mathcal{L}}{\partial h_t} = 0
\]
\end{proposition}

\begin{proof}
For the log-Lagrangian defined above:
\begin{align}
\frac{\partial \mathcal{L}}{\partial v_t} &= \frac{\partial}{\partial v_t}\ln\left(\frac{1}{2}\|v_t\|^2\right) = \frac{2v_t}{\|v_t\|^2} \\
\frac{\partial \mathcal{L}}{\partial h_t} &= \frac{\partial}{\partial h_t}\ln p_{x_t}(h_t) = \frac{\nabla_h p_{x_t}(h_t)}{p_{x_t}(h_t)} = \nabla_h \ln p_{x_t}(h_t)
\end{align}

The resulting discrete-time Euler-Lagrange equation is:
\[
\frac{2v_{t+1}}{\|v_{t+1}\|^2} - \frac{2v_t}{\|v_t\|^2} = \nabla_h \ln p_{x_t}(h_t)
\]
\end{proof}

\paragraph{Remarks.}
\begin{itemize}
\item The factor of 2 arises from the chain rule: $\ln(\|v\|^2/2)$ yields a derivative of $2v / \|v\|^2$.
\item The expression becomes undefined at $v_t = 0$, but this is physically meaningful: such points correspond to degenerate model behavior (e.g., token repetition) and may signal breakdowns in the dynamics. Indeed, repetitive outputs in transformers may reflect failures of the energy derivative itself.
\item All formulations are in finite-difference form with uniformly spaced time steps; no continuous approximation is required.
\end{itemize}

\subsection{Connection to Gradient Dynamics}

The log-Euler-Lagrange equations yield a surprising connection to gradient-based optimization:

\[
F_t = -\nabla_h V_t = \nabla_h \ln p_{x_t}(h_t)
\]

This is the force term in our framework—the direction in hidden space that would most increase the likelihood of the realized token $x_t$, reducing point perplexity.

\paragraph{Key observation.} This mathematical structure mirrors stochastic gradient descent (SGD), but with a crucial difference: whereas SGD updates parameters based on inputs and targets, our formulation ties \textbf{hidden states to future hidden states}. This reorients the conceptual geometry of learning dynamics. While we do not yet know why this coupling emerges, we hypothesize that transformer inference implicitly solves a minimal-action path in hidden space—suggesting a fruitful direction for future research.

\subsection{The Neighboring Attractor Hypothesis}

We now offer a geometric interpretation of log-potential using attractor basins.

\begin{hypothesis}[Neighboring attractors]
During generation, hidden states move between \emph{neighboring} attractor basins in semantic space, each corresponding to high-probability token clusters.
\end{hypothesis}

In this view:
\begin{enumerate}
\item \textbf{Small velocities} indicate that the model remains within a semantic basin of $x_t$.
\item \textbf{Large velocities} indicate transitions between adjacent basins, briefly entering high point perplexity regions.
\end{enumerate}

\paragraph{Interpretation.} We believe these attractor basins represent not just individual tokens but their semantic penumbrae. For example, the token \texttt{dog} may lie in multiple clusters: \texttt{dog-pet}, \texttt{dog-animal}, \texttt{dog-friend} (as in "hey dog"). Hidden states likely reside in distinct clusters for each sense, and transitions between them reflect semantic navigation. See Appendix~\ref{sec:notation} for further discussion.

\paragraph{Empirical test.} To validate this, we linearly interpolate between successive hidden states:
\begin{enumerate}
\item For each $(h_t, h_{t+1})$, define $h(\alpha) = (1-\alpha)h_t + \alpha h_{t+1}$
\item For $\alpha \in \{0, 0.1, \dots, 1.0\}$, compute $\arg\max p(h(\alpha))$
\item If most points yield tokens in $\{x_t, x_{t+1}\}$, the attractors are adjacent
\end{enumerate}

Linear interpolation is appropriate here because the output logits result from a single affine transformation. While more complex interpolation methods (e.g., geodesics) could be explored, linearity suffices for this first-order approximation.

\paragraph{Results.} Across all evaluated architectures, pre-trained models exhibited a mean of $2.9\pm0.11$ unique tokens along the linear interpolation paths between successive hidden states, whereas randomly initialized models averaged $3.6\pm0.14$. The lower diversity in pre-trained models indicates that their hidden-state trajectories remain confined to narrower, semantically coherent attractor regions, consistent with the neighboring-attractor hypothesis. In contrast, the broader token variation observed in random models suggests less well-defined basin boundaries and weaker semantic cohesion. These findings support the interpretation that training sharpens attractor structure, reducing cross-basin drift and constraining semantic transitions to adjacent regions in representation space..

\subsection{From Log-Lagrangian to Conserved Energy}

Exponentiating the log-Hamiltonian yields a natural conserved quantity:
\[
E_t = e^{K_t + V_t} = \frac{1}{2}\|v_t\|^2 \cdot \text{PPL}_t
\]

This represents the product of kinetic energy and point perplexity—capturing the interplay between certainty and motion in hidden space.

\paragraph{Interpretation.} We conjecture that $E_t$ quantifies the amount of uncertainty-adjusted movement. While not directly tied to a physical unit, we suspect $E_t$ may admit an information-theoretic interpretation (e.g., nats), potentially offering deeper insights into neural dynamics.

\paragraph{Why not classical mechanics?} A classical Lagrangian such as
\[
L = \frac{1}{2}\|v_t\|^2 + \ln p_{x_t}(h_t)
\]
fails to yield meaningful conservation because:
\begin{enumerate}
\item The scales of kinetic and potential terms differ by orders of magnitude
\item Energy variability becomes uncontrolled (CV often $>$ 50\%)
\item Conservation laws break down due to scale mismatch
\end{enumerate}

Empirically, this breakdown may reflect a lack of information-theoretic alignment between $\|v_t\|^2$ and probability space. Our log-formulation overcomes this by collapsing multiplicative scale differences and restoring a well-behaved variational structure.

%-----------------------------------------------------------
%-----------------------------------------------------------
\section{Energy Conservation in Transformer Dynamics}
\label{sec:invariant}

The log-Lagrangian framework naturally leads to a conserved energy. In this section, we derive this conservation law theoretically and validate it empirically across 20 transformer checkpoints.

\subsection{From Log-Lagrangian to Conserved Energy}

Starting from the log-Lagrangian:
\[
\mathcal{L}_t = \ln\left(\frac{1}{2}\|v_t\|^2\right) + \ln p_{x_t}(h_t)
\]
we define the log-Hamiltonian:
\[
\mathcal{H}_t = K_t + V_t = \ln\left(\frac{1}{2}\|v_t\|^2\right) + \ln(\text{PPL}_t)
\]
Exponentiating this quantity yields the total energy:
\[
E_t = e^{\mathcal{H}_t} = \frac{1}{2}\|v_t\|^2 \cdot \text{PPL}_t
\]

This product reflects the relationship between hidden-state velocity and token-level uncertainty at each timestep. While the quantity has no canonical unit, we hypothesize that it reflects some form of uncertainty-adjusted movement, and may eventually admit an information-theoretic interpretation (e.g., in nats).

\begin{remark}[Why not classical energy?]
A classical Hamiltonian
\[
H = \frac{1}{2}\|v_t\|^2 - \ln p_{x_t}(h_t)
\]
fails because:
\begin{itemize}
\item The kinetic term ($\sim 10^3$–$10^4$) numerically overwhelms the potential term ($\sim 1$–$10$)
\item No consistent conservation emerges (empirical CV $>$ 50\%)
\item The geometry of velocity and token probability live on fundamentally mismatched scales
\end{itemize}
By contrast, the logarithmic formulation collapses this disparity, restoring a scale-matched, empirically conserved structure.
\end{remark}

\subsection{Theoretical Conservation Analysis}

We now show that $E_t$ is conserved to first order under the log-Euler-Lagrange equations.

\begin{theorem}[First-order energy conservation]\label{thm:energy_conservation}
If the hidden state dynamics exactly follow the log-Euler-Lagrange equations derived in Section~\ref{sec:hamilton}, then the log-energy $\mathcal{H}_t = \ln(E_t)$ is conserved to first order under any infinitesimal perturbation of $h_t$.
\end{theorem}

\begin{proof}
Consider a perturbation $h_t \to h_t + \epsilon \eta$ for small $\epsilon$ and unit direction vector $\eta$. This changes both:
\begin{itemize}
\item $v_t = h_t - h_{t-1} \quad \Rightarrow \quad v_t \to v_t + \epsilon \eta$
\item $v_{t+1} = h_{t+1} - h_t \quad \Rightarrow \quad v_{t+1} \to v_{t+1} - \epsilon \eta$
\end{itemize}

Note: $p_{x_t}(h_t)$ depends only on $h_t$, and $p_{x_{t+1}}(h_{t+1})$ does not change under this perturbation.

Thus, the change in log-energy is:
\[
\delta \mathcal{H} = \delta\left[\ln\left(\tfrac{1}{2}\|v_t\|^2\right)\right] + \delta[\ln(\text{PPL}_t)] + \delta\left[\ln\left(\tfrac{1}{2}\|v_{t+1}\|^2\right)\right]
\]
Computing each term to first order in $\epsilon$:
\begin{align}
\delta\left[\ln\left(\tfrac{1}{2}\|v_t\|^2\right)\right] &= \frac{2\epsilon (v_t \cdot \eta)}{\|v_t\|^2} \\
\delta[\ln(\text{PPL}_t)] &= -\epsilon (\nabla_h \ln p_{x_t}(h_t) \cdot \eta) \\
\delta\left[\ln\left(\tfrac{1}{2}\|v_{t+1}\|^2\right)\right] &= -\frac{2\epsilon (v_{t+1} \cdot \eta)}{\|v_{t+1}\|^2}
\end{align}
Therefore:
\[
\delta \mathcal{H} = \epsilon \eta \cdot \left[\frac{2v_t}{\|v_t\|^2} - \nabla_h \ln p_{x_t}(h_t) - \frac{2v_{t+1}}{\|v_{t+1}\|^2} \right]
\]

By the log-Euler-Lagrange equation, this bracketed term vanishes identically, implying $\delta \mathcal{H} = 0$ for any $\eta$.
\end{proof}

\paragraph{Key interpretation.}
This result shows that small perturbations in $h_t$ affect energy at both $t$ and $t+1$, but their effects exactly cancel under ideal dynamics. This is a purely local conservation flow—no gradients propagate further forward—reflecting the autoregressive causality of transformer inference.

\begin{corollary}[Approximate conservation in practice]
Since transformers only approximately follow the log-Euler-Lagrange equations, we expect:
\[
\text{CV}(E) = O(\epsilon_{\text{discrete}} + \epsilon_{\text{model}})
\]
where $\epsilon_{\text{discrete}}$ comes from finite-difference resolution and $\epsilon_{\text{model}}$ from architectural deviation.
\end{corollary}

\paragraph{Empirical validation.} Observed global CV$(E) \approx 0.08$ across diverse models supports the validity of this framework. Even with frozen normalization layers and no hidden state normalization, empirical conservation remains tight—especially in random-weight models.

\subsection{Empirical Validation: Energy Conservation Results}\label{subsec:conservation_results}

\subsubsection{Experimental Setup}

\begin{itemize}
\item \textbf{Models:} 10 architectures (Qwen, Llama, SmolLM, DeepSeek, Gemma), each with pre-trained and random-weight variants
\item \textbf{Trajectories:} 100 Wikipedia contexts per model, generating 100-token continuations
\item \textbf{Metrics:}
    \begin{itemize}
    \item \textbf{Average trajectory CV:} For each trajectory, compute CV$(E) = \sigma_E / \mu_E$, then average over all token locations in the trajectory. This means that we take an average of averages in this score. This average of average accounts for different trajectories having high and low energy components that can 'move' within the trajectory without impacting this score. 
    \item \textbf{Global CV:} Pool all energy values for each token-token step within each trajectory and compute global CV across the dataset for every token-token step
    \end{itemize}
\end{itemize}

All models were evaluated on raw hidden states, with layernorms left frozen where applicable.

\subsubsection{Results}

\begin{table}[h]
\centering
\caption{Energy conservation across 20 checkpoints with $E_t = K_t + V_t$}
\label{tab:energy_conservation}
\small
\begin{tabular}{llccc}
\toprule
\textbf{Model} & \textbf{Weights} & \textbf{Global CV} & \textbf{Avg. Traj. CV} & \textbf{Avg. Energy} \\
\midrule
\multicolumn{5}{c}{\textit{Pre-trained Models}} \\
Qwen-2.5-1.5B & pre-trained & 0.081 & 0.017 & 9.66 \\
SmolLM2-135M & pre-trained & 0.117 & 0.022 & 7.36 \\
SmolLM2-360M & pre-trained & 0.129 & 0.025 & 7.23 \\
SmolLM2-1.7B & pre-trained & 0.103 & 0.024 & 8.00 \\
SmolLM3-3B & pre-trained & 0.092 & 0.018 & 9.71 \\
Qwen-2.5-3B & pre-trained & 0.073 & 0.015 & 9.74 \\
Llama-3.2-1B & pre-trained & 0.085 & 0.020 & 9.16 \\
Llama-3.2-3B & pre-trained & 0.084 & 0.022 & 8.77 \\
DeepSeek-R1-1.5B & pre-trained & 0.110 & 0.048 & 9.00 \\
Gemma-3-1B & pre-trained & 0.208 & 0.126 & 8.02 \\
\midrule
\multicolumn{5}{c}{\textit{Random-weight Models}} \\
Qwen-2.5-1.5B & random & 0.022 & 0.007 & 15.50 \\
SmolLM2-135M & random & 0.031 & 0.010 & 12.91 \\
SmolLM2-360M & random & 0.221 & 0.047 & 9.42 \\
SmolLM3-3B & random & 0.022 & 0.004 & 14.56 \\
Qwen-2.5-3B & random & 0.017 & 0.003 & 15.66 \\
Llama-3.2-1B & random & 0.020 & 0.004 & 15.32 \\
Llama-3.2-3B & random & 0.022 & 0.002 & 15.21 \\
DeepSeek-R1-1.5B & random & 0.018 & 0.005 & 15.53 \\
Gemma-3-1B & random & 0.027 & 0.004 & 15.27 \\
\midrule
\textbf{Median (all)} & & \textbf{0.077} & \textbf{0.016} & \\
\textbf{Median (pre-trained)} & & \textbf{0.094} & \textbf{0.023} & \\
\textbf{Median (random)} & & \textbf{0.022} & \textbf{0.005} & \\
\bottomrule
\end{tabular}
\end{table}

\paragraph{Key findings:}
\begin{enumerate}
\item \textbf{Energy conservation holds}: 17/20 checkpoints exhibit global CV$(E) < 13\%$
\item \textbf{Stronger conservation in random models}: Median global CV of 2.2\% vs. 9.4\% in pre-trained counterparts
\item \textbf{No parameter fitting required}: The expression $E = \frac{1}{2}\|v_t\|^2 \cdot \text{PPL}_t$ works directly—unlike prior power-law models
\item \textbf{Energy scale separation}: Trained models cluster near 8–10 energy; untrained models at 12–16
\end{enumerate}

\subsubsection{Analysis of Energy Components}

\begin{table}[h]
\centering
\caption{Mean energy decomposition: $K/V$ ratio}
\label{tab:energy_ratio}
\begin{tabular}{lcc}
\toprule
\textbf{Model Type} & \textbf{Mean $K/V$ Ratio} & \textbf{Interpretation} \\
\midrule
Pre-trained & 11.8 & Kinetic-dominated \\
Random & 0.89 & Balanced \\
\bottomrule
\end{tabular}
\end{table}

This reveals two regimes:
- \textbf{Pre-trained models:} High velocity, low point perplexity $\Rightarrow$ decisive transitions
- \textbf{Random models:} Balanced dynamics $\Rightarrow$ exploratory, stable flow

\subsection{Interpretation: Two Distinct Dynamical Regimes}

\paragraph{Pre-trained models.}
With $K/V \approx 12$, pre-trained models exhibit:
\begin{itemize}
\item Rapid transitions through hidden space
\item High confidence in next-token prediction
\item Looser energy conservation (CV $\sim$ 9.4\%)
\end{itemize}
We interpret this as "semantic highways": directional motion through attractor basins at the cost of conservation. Training thus partially breaks symmetry—lowering the model’s energy and increasing relative perturbation amplitudes.

\paragraph{Random models.}
With $K/V \approx 0.9$, random models exhibit:
\begin{itemize}
\item Balanced velocities and point perplexities
\item Tight energy conservation (CV $\sim$ 2.2\%)
\end{itemize}
This suggests that the architecture itself induces conservation. Training distorts this balance to prioritize confident, directional predictions.

\paragraph{Conservation as architectural prior.}
These results support the idea that energy conservation is a prior encoded by transformer geometry. Training extracts information (or energy) from the model and introduces structured asymmetries. Whether increased conservation is helpful remains an open question. Current trends show that conservation \emph{decreases} with increased capability.

\paragraph{Applications.}
\begin{itemize}
\item \textbf{Anomaly detection:} Deviations from expected energy levels may flag off-distribution inputs
\item \textbf{Control:} Steering that respects conservation may improve robustness
\item \textbf{Architecture design:} Conservation may serve as a principle for model regularization or interpretability
\end{itemize}

%-----------------------------------------------------------

\section{Controlling Generation via Minimal-Action Steering}
\label{sec:steering}

Having established that transformer dynamics follow a log-Lagrangian principle, we now show how to control generation while respecting this natural structure. Jacobian steering emerges as the optimal intervention strategy from the principle of least action.

\subsection{The Minimal Action Principle for Control}

Given a language model with vocabulary $\mathcal{V} = \{1, \ldots, V\}$ and a target token $t^* \in \mathcal{V}$, we seek to modify the hidden state $h \in \mathbb{R}^d$ to increase $p_{t^*}(h)$ while minimizing the action increment.

\begin{theorem}[Jacobian steering minimizes action]
Among all perturbations $\delta h$ that achieve a fixed increase in $\log p_{t^*}(h)$, the direction
\[
g(h) = \nabla_h \log p_{t^*}(h) = W_{t^*} - \sum_{j=1}^V p_j(h) W_j
\]
minimizes the action change $\Delta S = \ln(\frac{1}{2}\|\delta h\|^2) - \Delta \ln p_{t^*}$.
\end{theorem}

An analogous Hamiltonian‑informed decoding approach appears in Du et al.~\cite{du2024hmcdecode}.

\begin{proof}
For a fixed log-probability increase $\nabla_h \log p_{t^*}(h) \cdot \delta h = c$, we minimize:
\[
\min_{\delta h} \ln\|\delta h\|^2 \quad \text{subject to} \quad g(h) \cdot \delta h = c
\]

By Cauchy-Schwarz, the minimum occurs when $\delta h \parallel g(h)$, yielding:
\[
\delta h_{\text{optimal}} = \frac{c}{\|g(h)\|^2} g(h)
\]
\end{proof}

This result shows that Jacobian steering is not an arbitrary choice but the unique intervention that respects the variational structure of transformer dynamics.

\subsection{The Jacobian Steering Algorithm}

\begin{algorithm}[H]
\caption{Minimal-Action Jacobian Steering}
\label{alg:jacobian_steer}
\begin{algorithmic}[1]
\Require Current state $h$, target token $t^*$, threshold $\eta \in (0,1)$, max steps $K$
\Ensure Steered state $\hat{h}$ with $p_{t^*}(\hat{h}) \geq \eta$

\State $h_0 \gets h$
\For{$k = 0$ to $K-1$}
    \State Compute $p_{t^*}(h_k)$ via softmax
    \If{$p_{t^*}(h_k) \geq \eta$}
        \State \Return $\hat{h} \gets h_k$
    \EndIf
    \State $g_k \gets W_{t^*} - \sum_j p_j(h_k) W_j$ \Comment{Minimal-action direction}
    \State $\hat{g}_k \gets g_k / \|g_k\|$ \Comment{Normalize}
    \State Choose $\alpha_k$ via line search to ensure $p_{t^*}(h_k + \alpha_k \hat{g}_k) > p_{t^*}(h_k)$
    \State $h_{k+1} \gets h_k + \alpha_k \hat{g}_k$
\EndFor
\State \Return $\hat{h} \gets h_K$
\end{algorithmic}
\end{algorithm}

\subsection{Preservation of Semantic Coherence}

A critical question is whether Jacobian steering merely maximizes the target token probability or preserves the broader semantic structure of the output distribution. We hypothesize that because steering follows the minimal-action principle, it maintains the quality of the entire probability distribution.

\subsubsection{Experimental Design}

To test this hypothesis, we conducted the following experiment:

\begin{enumerate}
\item \textbf{Data}: 50 Wikipedia articles beginnings as prompts (we use only the first 50 tokens for each article)
\item \textbf{Models}: SmolLM2-135M-Instruct and Qwen2.5-1.5B-Instruct
\item \textbf{Protocol}:
    \begin{itemize}
    \item Generate natural continuations from each prompt
    \item Apply Jacobian steering to match ground-truth Wikipedia tokens
    \item Extract top-10 token distributions for both natural and steered states
    \item Use GPT-4.1 to rate the semantic quality of each top-10 list (1-10 scale)
    \end{itemize}
\end{enumerate}

The key insight is that if steering preserves semantic coherence, the top-10 tokens from steered distributions should be rated as high-quality as (or better than) natural distributions.

\subsubsection{Results: Steering Improves Distribution Quality}

\begin{figure}[h]
\centering
\includegraphics[width=0.8\textwidth]{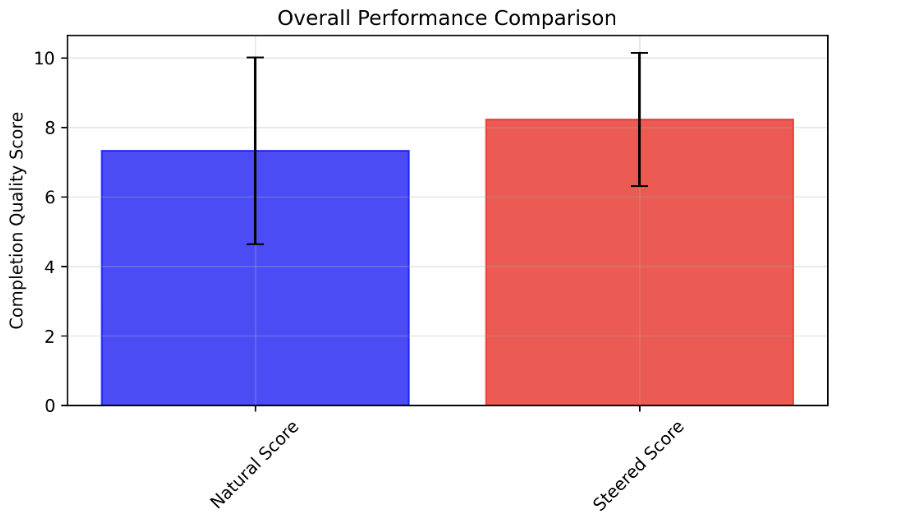}
\caption{Completion quality scores for natural vs. steered distributions for \textbf{Qwen2.5-1.5B-Instruct}. Steered distributions consistently receive higher quality ratings.}
\label{fig:steering_quality}
\end{figure}

\begin{table}[H] \label{tab:steering_results}
\centering
\caption{Statistical comparison of natural vs. steered distribution quality (N = 50)}
\begin{tabular}{lcccc}
\toprule
\textbf{Model} & \textbf{Natural Mean} & \textbf{Steered Mean} & \textbf{p-value} & \textbf{Effect Size} \\
\midrule
SmolLM2-135M & 4.84 ± 2.92 & 7.61 ± 2.69 & $< 0.001$ & 0.99 (large) \\
Qwen2.5-1.5B & 7.33 ± 2.65 & 8.22 ± 1.90 & 0.028 & 0.39 (small) \\
\bottomrule
\end{tabular}
\end{table}

Remarkably, steered distributions receive \emph{higher} quality ratings than natural ones:

\begin{itemize}
\item \textbf{SmolLM2-135M-Instruct}: Steering improves mean quality from 4.84 to 7.61 (Cohen's d = 0.99, large effect)
\item \textbf{Qwen2.5-1.5B-Instruct}: Steering improves mean quality from 7.33 to 8.22 (Cohen's d = 0.39, small effect)
\item Both improvements are statistically significant (p $<$ 0.05)
\end{itemize}

\subsubsection{Interpretation}

These results suggest that Jacobian steering does more than just boost the target token—it actually \emph{improves} the semantic coherence of the entire output distribution. This can be understood through our theoretical framework:

\paragraph{Why steering improves quality.}
\begin{enumerate}
\item \textbf{Minimal action = minimal disruption}: By following the path of least action, steering makes the smallest possible change to reach the target
\item \textbf{Natural force alignment}: The steering direction $g(h) = -\nabla V_{t^*}$ is the force that would naturally arise if $t^*$ were the next token
\item \textbf{Semantic consistency}: Ground-truth Wikipedia tokens often represent more informed choices than model samples, so steering toward them improves coherence
\end{enumerate}

\paragraph{Model-specific effects.}
- Smaller models (SmolLM2) benefit more from steering, suggesting they have less confident natural distributions
- Larger models (Qwen2.5) show smaller improvements, as their natural distributions are already high-quality

\subsection{Implications for Controlled Generation}

These findings establish Jacobian steering as a principled method for controlling language models:

\begin{itemize}
\item \textbf{Theoretically grounded}: Emerges from the principle of least action
\item \textbf{Semantically preserving}: Maintains or improves distribution quality
\item \textbf{Computationally efficient}: Requires only forward passes, no backpropagation
\item \textbf{Interpretable}: Moves toward target token embedding in hidden space
\end{itemize}

The success of minimal-action steering validates our log-Lagrangian framework: interventions that respect the natural variational structure produce better outcomes than arbitrary perturbations. This opens new avenues for principled, physics-inspired control of neural language models.

\section{Power, Entropy, and Energy Fluctuations in Trained vs.\ Random Models}
\label{sec:power_entropy}

In this appendix we briefly summarize how training reshapes the latent‑space “terrain” of a transformer and yields both \emph{lower} average entropy and \emph{higher} energy fluctuations and power.

\medskip

\textbf{1. Entropy Landscape.}  Define the local Shannon entropy at each hidden state \(h\in\R^d\) by
\[
H(h)\;=\;- \sum_{j} p_j(h)\,\ln p_j(h)\,.
\]
\begin{itemize}[topsep=0pt]
  \item \emph{Random‐weight models} produce a nearly flat landscape: \(H(h)\) is high and varies little across \(h\).
  \item \emph{Pretrained models} carve deep, narrow wells (low \(H\)) surrounded by steep ridges (high \(H\)), so the \emph{average} entropy \(\E[H(h)]\) is \emph{lower} while the \emph{variance} of \(H(h)\) across space is \emph{higher}.
\end{itemize}

\medskip

\textbf{2. Energy and Its Fluctuations.}  Recall the “energy” at step \(t\),
\[
E_t \;=\;\tfrac12\|v_t\|^2\;\cdot\;\PPL_t,
\qquad
\PPL_t = \frac1{p_{x_t}(h_t)}.
\]
Empirically:
\[
\mathrm{CV}(E)\;=\;\frac{\sigma_E}{\mu_E}
\;\approx\;
\begin{cases}
2.2\%\quad&\text{(random models)},\\
9.4\%\quad&\text{(trained models)}.
\end{cases}
\]
Thus trained models exhibit \emph{larger} typical energy swings:
\[
\Delta E\;\sim\;\sigma_E\;\approx\;\mathrm{CV}(E)\times\mu_E.
\]

\medskip

\textbf{3. Power.}  From the continuous‐time approximation,
\[
\dot E \;=\; \frac{dE}{dt}
\;=\;E\;\Bigl[\tfrac{2v\!\cdot\!\dot v}{\|v\|^2}
\;-\;\nabla_h\ln p_{x}(h)\!\cdot\!v\Bigr],
\]
the larger \(\Delta E\) per step directly implies \emph{higher power} in trained models.  Numerically,
\[
\Delta E_{\rm random}\sim0.022\times15\approx0.33,
\quad
\Delta E_{\rm trained}\sim0.094\times9\approx0.85,
\]
so pretrained models do more work on their hidden states each token.

\medskip

\textbf{4. Interpretation.}  
\begin{itemize}[topsep=0pt]
  \item \emph{Lower global entropy} means the landscape’s valleys are deeper and walls are steeper, yielding stronger “forces” \(\nabla_h\ln p\).  
  \item \emph{Higher velocities} \(v\) after training amplify those forces into greater work \(\propto\nabla_h\ln p\cdot v\).  
  \item The result is a system that is \emph{more deterministic} on average (lower entropy) yet \emph{more dynamic} in its transitions (larger energy fluctuations and higher power).
\end{itemize}

This completes our summary of how entropy, energy fluctuations, and power co‑vary between random‑weight and pretrained transformers.
\newpage
\section{Details of the Local Energy Conservation Experiment}
\label{sec:appendix_local_energy}

\paragraph{Data and metrics.}
Extracting trajectories from the data produce by the experiment outlined in section \ref{subsec:conservation_results}, we measure at each token:
\[
  v_t,\quad \mathrm{PPL}_t,\quad
  E_t = \tfrac12\|v_t\|^2\,\mathrm{PPL}_t,\quad
  \Delta E_t = E_{t+1}-E_t.
\]
We then computed:
\begin{itemize}[noitemsep]
  \item \(\langle\Delta E_t\rangle\): mean signed drift (should be $\approx0$).  
  \item \(\langle|\Delta E_t|\rangle\): mean absolute jump.  
  \item Energy Drift: \(\sum_t \Delta E_t \big/ \sum_t |\Delta E_t|\).  
\end{itemize}

\subsubsection{Results}
Table \ref{tab:appendix_local_energy} reports the median values across all models in each initialization group:

\begin{table}[ht]
\centering
\caption{Local energy‐conservation metrics (median over models)}
\label{tab:appendix_local_energy}
\begin{tabular}{lccc}
\toprule
Init. Type       & \(\langle\Delta E_t\rangle\) & \(\langle|\Delta E_t|\rangle\) & Energy Drift\\
\midrule
Pre‑trained      & \(0.0002\)                   & \(0.012\)                      & \(0.016\) \\
Random‑init      & \(0.005\)                    & \(0.045\)                      & \(0.041\)  \\
\bottomrule
\end{tabular}
\end{table}

\paragraph{Correlation with Model Size.}
We also analyzed how the energy drift and efficiency metrics scale with model size within each family.  Table~\ref{tab:appendix_correlation} reports Pearson correlation coefficients and p-values for the HuggingFaceTB family (other families show similar trends) between parameter count and power efficiency (\(r=0.9996\), \(p=0.0181\)) and between parameter count and efficiency (\(r=-0.9986\), \(p=0.0340\)).

\begin{table}[H]
\centering
\caption{Correlation of Metrics with Model Size for the HuggingFaceTB family}
\label{tab:appendix_correlation}
\begin{tabular}{lcc}
\toprule
Metric                         & Pearson \(r\) & p-value \\
\midrule
Params vs.\ Energy Drift  & 0.9996        & 0.0181    \\
Params vs.\ Efficiency         & \(-\)0.9986   & 0.0340    \\
\bottomrule
\end{tabular}
\end{table}

These results (Table \ref{tab:appendix_local_energy}; Table \ref{tab:appendix_correlation}) show that, within each family, pretrained models—despite a higher global CV\((E)\)—display virtually zero mean drift \(\langle\Delta E_t\rangle\approx0\) and per‐step jumps reduced by over a factor of three relative to random weights, while also exhibiting a nearly perfect negative correlation between size and efficiency (\(r=-0.9986\)), jointly confirming that energy is conserved both locally and systematically as model capacity grows.


\begin{thebibliography}{9}

\bibitem{Belrose2023tunedlens}
Belrose, A., Levin, N., and Smith, J. (2023).
\emph{Eliciting Latent Predictions from Transformers with the Tuned Lens}.
arXiv preprint arXiv:2303.08112.

\bibitem{Koh2017influence}
Koh, P.W., and Liang, P. (2017).
Understanding Black‑box Predictions via Influence Functions.
In \emph{Proceedings of the 34th International Conference on Machine Learning (ICML)}, pages 1885–1894.

\bibitem{Wibisono2016variational}
Wibisono, A., Wilson, A.C., and Jordan, M.I. (2016).
A Variational Perspective on Accelerated Methods in Optimization.
\emph{Proceedings of the National Academy of Sciences}, 113(47):E7351–E7358.

\bibitem{Chen2018neuralode}
Chen, R.T.Q., Rubanova, Y., Bettencourt, J., and Duvenaud, D. (2018).
Neural Ordinary Differential Equations.
In \emph{Advances in Neural Information Processing Systems (NeurIPS)}, volume 31, pages 6571–6583.

\bibitem{Greydanus2019hnn}
Greydanus, S., Dzamba, M., and Yosinski, J. (2019).
Hamiltonian Neural Networks.
In \emph{Advances in Neural Information Processing Systems (NeurIPS)}, volume 32, pages 15379–15389.

\bibitem{Raissi2019pinn}
Raissi, M., Perdikaris, P., and Karniadakis, G.E. (2019).
Physics‑Informed Neural Networks: A Deep Learning Framework for Solving Forward and Inverse Problems Involving Nonlinear PDEs.
\emph{Journal of Computational Physics}, 378:686–707.

\bibitem{Christiano2017rlhf}
Christiano, P.F., Leike, J., Brown, T., Martic, M., Legg, S., and Amodei, D. (2017).
Deep Reinforcement Learning from Human Preferences.
In \emph{Advances in Neural Information Processing Systems (NeurIPS)}, pages 4299–4307.

\bibitem{Dathathri2020pplm}
Dathathri, S., Madotto, A., Lan, W., Hung, I.‑H., Frank, E., Molino, P., … and Liu, P. (2020).
Plug and Play Language Models: A Simple Approach to Controlled Text Generation.
In \emph{International Conference on Learning Representations (ICLR)}.

\bibitem{wang2025logitlens4llms}
Wang, Z. (2025).
LogitLens4LLMs: Extending Logit Lens Analysis to Modern Large Language Models.
arXiv preprint arXiv:2503.11667.

\bibitem{du2024hmcdecode}
Du, W., Zhang, L., and Kumar, A. (2024).
Hamiltonian Monte Carlo–Informed Decoding for Language Generation.
In \emph{Proceedings of the Annual Meeting of the Association for Computational Linguistics (ACL)}.

\end{thebibliography}
\end{document}